# ALARMS: Alerting and Reasoning Management System for Next Generation Aircraft Hazards

**Alan S. Carlin** *
University of Massachusetts
Computer Science Building
140 Governors Drive
Amherst, MA 01003
acarlin@cs.umass.edu

**Nathan Schurr**
Aptima, Inc.
12 Gill Street
Suite 1400
Woburn, MA 01801
nschurr@aptima.com

**Janusz Marecki**
IBM T.J Watson Research
1101 Kirchawan Road, Route 134
Yorktown Heights, NY 10598
marecki@us.ibm.com

## Abstract

The Next Generation Air Transportation System will introduce new, advanced sensor technologies into the cockpit. With the introduction of such systems, the responsibilities of the pilot are expected to dramatically increase. In the ALARMS (ALerting And Reasoning Management System) project for NASA, we focus on a key challenge of this environment, the quick and efficient handling of aircraft sensor alerts. It is infeasible to alert the pilot on the state of all subsystems at all times. Furthermore, there is uncertainty as to the true hazard state despite the evidence of the alerts, and there is uncertainty as to the effect and duration of actions taken to address these alerts.

This paper reports on the first steps in the construction of an application designed to handle Next Generation alerts. In ALARMS, we have identified 60 different aircraft subsystems and 20 different underlying hazards. In this paper, we show how a Bayesian network can be used to derive the state of the underlying hazards, based on the sensor input. Then, we propose a framework whereby an automated system can plan to address these hazards in cooperation with the pilot, using a Time-Dependent Markov Process (TMDP). Different hazards and pilot states will call for different alerting automation plans. We demonstrate this emerging application of Bayesian networks and TMDPs to cockpit automation, for a use case where a small number of hazards are present, and analyze the resulting alerting automation policies.

## 1 Introduction

Next Generation Air Transportation System technologies will introduce new, advanced sensor technologies into the cockpit. With the introduction of such systems, the responsibilities of the pilot and the density of air traffic are both expected to dramatically increase (Joint Planning and Development Office, 2007). As a result, the number of potential hazards and relevant information that must be perceived and processed by the pilot will grow. This information is likely to come from a variety of sources, requiring the pilot to integrate this information in order to evaluate hazard potential. Evaluating hazard potential will depend on the consideration of, and differentiation between, immediate (current) hazards and situations requiring re-planning or coordination (future). It will also require reasoning under uncertainty, as the actual state of the world needs to be reasoned from the hazards, and also a plan needs to be constructed for the pilot and artificial aircraft intelligence to handle the hazards, despite uncertainty as to the effectiveness of each, and temporal uncertainty about the duration required to handle each hazard.

To support these challenging responsibilities, the pilot has an unambiguous need for an Integrated Alerting and Notification (IAN) system that will continuously monitor multiple sources of interdependent information to evaluate hazard potentials, track multiple potential hazards, provide caution/warning/alerting (CWA) notifications and context-relevant decision support to the pilot, and determine the best method of presenting this information to ensure that the information can be viewed and used efficiently and effectively. There are two broad challenges that need to be addressed before an IAN system can become operational. First, existing methods cannot reason under uncertainty about the proposed scale of information and hazards in such a time-critical environment (Proctor, 1998; Song and Kuchar, 2003). Specifically, these methods do not provide a robust approach for integrating, interpreting, and providing recommendations that can be generated by the diverse and large set of data expected within the NextGen concept of operations. Second, the interaction between a human pilot and an automated system is complex (Galster, 2003) and also uncertain, and the design of an advanced alerting technology must leverage methods for ensuring effective collaborative performance of the human-system team.

---
*This author conducted this work while at Aptima, Inc.

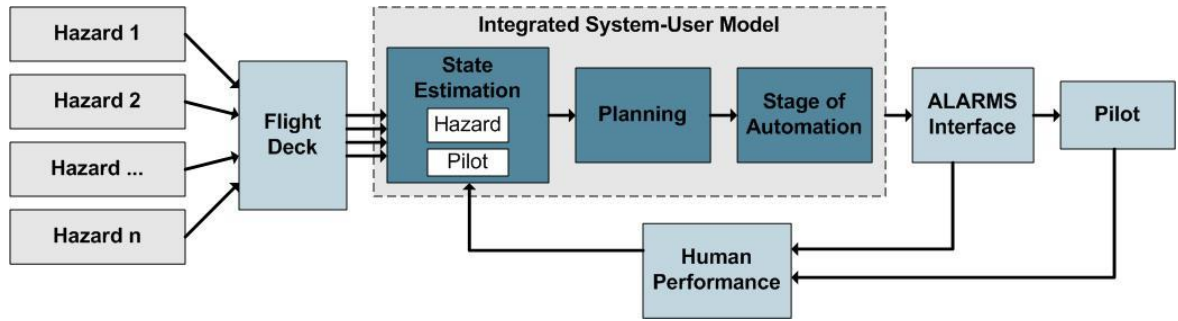

Figure 1: The ALARMS approach. Hazards in the environment are detected by aircraft sensors, and pilot state is estimated through Cognitive Work Analysis in the Human Performance module. A Bayesian Network (State Estimation) weights the sensor output to estimate the hazard state. A Planning module forms a time-sensitive plan for the pilot and automated system to address these hazards. The result is a plan with various stages of automation. The ALARMS Interface displays information at the appropriate stage to the pilot.

## 2 ALARMS Approach

To address these challenges, we have designed and simulated an ALerting And Reasoning Management System (ALARMS). The ALARMS approach is shown in Figure 1. Hazards exist in the real world, as depicted on the left of the diagram. The sensors on the Flight Deck (current and NextGen) perceive these hazards. In the first phase of the ALARMS effort, Aptima cooperated with the Aircraft Simulation and Hardware-in-the-Loop Lab to identify the hazards and sensor systems, and quantify the relationship between them. The results of this effort will be described in the next section. We used the results of this analysis to construct probability tables for a State Estimation Bayesian Network, labeled "State Estimation" in the Figure 1. The input to the State Estimation Bayesian Network is the alerts issued by the sensor systems on the aircraft. The output is an estimate of the probability and severity estimate of the underlying hazards. Once the hazard and pilot state is estimated, the information is sent to the ALARMS Planning module, which will construct a plan to address the hazards. The Planning module is a TMDP (Time-Dependent Markov Decision Processes) (Boyan and Littman, 2000). The TMDP model can be used to capture both state uncertainty in the environment as well as duration uncertainty in human (pilot) actions (Schurr and Marecki, 2008). Its input is the hazard and pilot states, as well as a Markov model of the effectiveness of the pilot and automation in handling the hazards, given various levels of alert. Its output is a time-dependent plan for addressing the alert. The plan is interpreted by the Stages of Automation module, which interprets the level of automation and decides what level of alerts and options to send to the pilot. This decision will then be sent to the ALARMS interface, which displays the information to the pilot.

In this paper, we report on the phase of the ALARMS effort corresponding to the "Integrated System User Model" in Figure 1. The Integrated System User Model is part of the greater ALARMS effort. Other parts of the ALARMS effort are shown in Figure 1 as well, and involve modeling and predicting human performance. Through Cognitive Work Analysis (CWA) (Vicente, 1999), ALARMS has focused on allowing the system to predict pilot performance in an online fashion. This module is labeled "Human Performance" in the figure. Whereas we will see that the current work involves a primitive model of pilot state which accounts for the phase of flight, in future work the Pilot State Estimate will be constructed using techniques from the cognitive sciences literature, resulting in a richer pilot state space for the Integrated System User Model (Latorella, 1999). Another, orthogonal task, is labeled "ALARMS Interface" in Figure 1. In this task, we are leveraging established human factors design principles to develop an interface that (1) maximizes pilot performance in detecting and responding to a diverse set of threats, and (2) is flexible in its integration with existing NextGen features and the concepts of operations. The interface is being designed to support various stages of automation (Galster, 2003). At low stages, decisions are made almost entirely by the pilot, whereas at high-stages of automation they are made by the system. At medium stages of automation, decisions are cooperative, for example the system may pre-compute several options for complex maneuvers, and let the pilot select from these options.

### 2.1 ALARMS Hazard Matrix

As the first step in the ALARMS effort, we combined with the Aircraft Simulation and Hardware-in-the-Loop Laboratory to identify (1) Current and Next Generation aircraft systems. (2) Aircraft Hazards. (3) The interaction between systems and hazards. The ALARMS hazard matrix was constructed as a result of this work, and a small portion of this matrix is shown in Figure 3. Each row represents a tool, technology, or system that can issue an alert in the next generation cockpit. Each column represents a potential environment in the hazard. Thus, there is an entry for each case where a sensor can issue an alert for a given hazard.

| Alert | Timeframe |
|---|---|
| Directive | <10 seconds |
| Warning | 10--15 seconds |
| Caution | < 40 seconds |
| Advisory | non-critical |

Figure 2: Color-coded entries of the hazard matrix represent the level of alert, which in turn corresponds to the timeframe in which that alert must be addressed.

Entries also color-coded with labels D/W/C/A (Directive/Warning/Caution/Advisory). These labels, presented in order of decreasing urgency, represent the highest level of alert possible for the sensor/hazard combination. Thus, entries labeled "A" correspond to systems that will only issue low level advisories for that given hazard, whereas entries labeled "W", such as an Adverse Weather hazard from the Enhanced Ground Proximity Warning System (EGPWS), correspond to systems that are capable of issuing a more urgent warning. Alert level also corresponds to the timeframe in which the hazard must be addressed, according to Figure 2.

## 2.2 ALARMS Bayesian Network

The goal of the ALARMS effort is to construct a plan for handling sensor alerts; however, it is not truly the sensor alerts that must be handled, it is the underlying hazards which they represent. For example, multiple subsystems (such as Vertical Navigation (VNAV) and Cockpit Display of Traffic Information (CDTI)) can issue alerts which relate to an Altitude Deviation hazard. The ALARMS system should deduce that there may be an Altitude Deviation hazard if either of these sensors issue an alert, and the hazard should be more certain and urgent if both systems issue one.

In order to model the sensor systems, a Bayesian network was built, as shown in Figure 4. Ovals on the right represent hazards, corresponding to the columns of the ALARMS hazard matrix. Ovals on the left represent hazard alerts from sensors, corresponding to the entries in the hazard matrix. Directionality proceeds from right to left, indicating that hazards cause sensor alerts. However reasoning proceeds from left to right, the sensor output will act as evidence and the hazard level on the right is deduced. Each entry on both sides can exist at several levels (Advisory, Caution, Warning, Directive), according to the severity of the hazard and the sensor alert.

The Bayesian network itself is modeled as noisy-or. The hazard itself can exist at several levels, and the level or severity of the hazard is evidenced by the subsystem alerts. To feed into the next part of the system, a threshold is applied, to find the highest level alert for which there is evidence.

To construct the network, the GeNIE (Graphical Network Interface) and SMILE (Structural Modeling, Inference, and Learning Engine) software packages were used, from the Decision Systems Laboratory at the University of Pittsburgh (Druzdzel, 1999).

## 2.3 TMDP Planner

We now recall the TMDP model and then show how the ALARMS planning module employs it. Time Dependent Markov Decision Processes (TMDPs) (Boyan and Littman, 2000) assume a finite set $S$ of discrete states and a finite set $A$ of actions. When action $a \in A$ is executed in state $s \in S$, the process transitions with probability $P^{s,a}(s')$ to some state $s' \in S$. The transition itself is not instantaneous; it consumes $t$ units of time with probability $d_{s'}^{s;a}(t)$ where $d_{s'}^{s;a} \in D$ is a probability density function (a.k.a. *action duration distribution*) for a given $s, a, s'$. Similarly, the reward $R_{s'}^{s;a}(t)$ that the transition provides depends on $s, a, s'$ as well as on the time $t$ at which the process enters state $s'$. (Note, that $t$ in $d_{s'}^{s;a}(t)$ is the transition duration whereas $t$ in $R_{s'}^{s;a}(t)$ is the time at which the transition terminates.) A deterministic TMDP policy $\pi$ is therefore a mapping $S \times [0, \Delta] \to A$ where $\Delta$ (a.k.a. *the deadline*) is the earliest point in time after which all the rewards $R_{s'}^{s;a}(t)$ are zero. Denote by $V^\pi(s, t)$ the total expected reward for following a policy $\pi$ from state $s$ at time $t$. (For a given $s$, $V^\pi(s, t)$ is often viewed as a continuous *value function* over $t \in [0, \Delta]$.) The optimal TMDP policy $\pi^*$ thus satisfies $V^{\pi^*}(s, t) \geq V^\pi(s, t)$ for all $s \in S$, $t \in [0, \Delta]$ and $\pi \neq \pi^*$.

Let $\Psi$ be the set of alert levels (e.g. "Nominal" (N), "Advisory" (A), "Caution" (C), "Warning" (W), or "Directive" (D)), $\Phi$ be an ordered set of hazards (e.g. "Weather" (hazard 1), "Altitude Deviation" (hazard 2)) and $\Omega$ be a set of autonomy levels (e.g. "No Autonomy" (0), "Some Autonomy" (1), or "Full Autonomy" (2)). A TMDP in ALARMS planing module is instantiated as follows:

- **States**: A state $s \in S$ is a mapping from the hazards to their alert levels. That is, $s = (\psi_\phi)_{\phi \in \Phi}$ is a vector where $\psi_\phi \in \Psi$ is the alert level of hazard $\phi \in \Phi$. For example, given three hazards, state $s = (N, A, W)$ defines that the first hazard is at Nominal level, the second hazard is at Advisory level, and the third haz-ard is at Warning Level.

| Tools/Technology/Systems | Sub-Systems | System Failure | System Performance Compromised | Loss of Separation | Adverse Weather Encounter | Altitude Deviation |
|---|---|---|---|---|---|---|
| Electrical (3.1.11) | | (C) | (W) | | | |
| Hydraulic (3.1.15) | | (C) | (C) | | | |
| Fuel (3.1.14) | | (C) | (C) | | | |
| Landing Gear (3.1.17) | | (W) | (C) | | | |
| Air Conditioning/Pressurization (3.1.21/22) | | (W) | (C) | | | |
| Ice Protection (3.1.16) | | (W) | (W) | | (C) | |
| Fire Protection (3.1.12) | | (C) | (C) | | | |
| Enhanced Ground Proximity Warning (EGPWS) (3.1.20) | | (A) | (A) | | (W) | |
| Navigation Radio (Nav Radio) | | | | | | |
| | Enroute (3.1.1) | (A) | (A) | | | |
| | Approach (3.1.1) | (A) | (A) | | | (A) |
| Flight Management System (FMS) | | | | | | |
| | RNAV (3.1.18) | (A) | (A) | | | |

Figure 3: A portion of the ALARMS hazard matrix. Each row represents a different sensor subsystem (blue rows are aviation systems, green rows are navigation systems), each column after the Hazards label represents a potential aircraft hazard. Color-coded entries represent the highest urgency alert that the sensor system may issue.

- **Actions**: The actions of the ALARMS system represent the different ways in which the system displays the information about the hazards on the pilot's GUI. In general, the higher the degree of autonomy $\omega_\phi \in \Omega$ for a hazard $\phi \in \Phi$, the less intrusive the way in which the information about hazard $\phi$ is presented on the pilot's GUI. An action $a \in A$ is therefore represented by a vector $(\omega_\phi)_{\phi \in \Phi}$. For example, given three hazards, action $a = (1, 3, 2)$ will mark on the pilot's GUI the information about hazard 1 to be pilot-intensive (autonomy level 1), the information about hazard 2 to be highly automated (autonomy level 3) and the information about hazard 3 to be somewhat automated (autonomy level 2). There is also a special autonomy level 0 reserved for actions that do not address the hazard at all (the pilot is not informed about a hazard and the automation does not address it).

- **Transitions**: ALARMS assumes that all the hazards will eventually be addressed (their alert levels will return to "Nominal" (N) values) as a result of human or autonomy actions. That is, for all states $s \in S$ and actions $a \in A$, $P^{s,a}(s') = 1$ only for $s' = (\psi_\phi)_{\phi \in \Phi}$ such that $\psi_\phi = N$ for all $\phi \in \Phi$. An exception to the above is when the action is not to address the hazard, in which case the state remains the same.

- **Durations**: ALARMS models action duration distributions by assuming that actions at a high level of automation take place quickly whereas actions at a low level of automation (i.e. that involve the pilot) have a longer duration. In essence, an attentive pilot will be more efficient at addressing hazards whereas an inattentive or overburdened pilot will perform poorly. As part of the ALARMS effort, profiles of pilot performance have been constructed at various phases of flight. The phase of flight affects pilot attentiveness, which in turn affects the action duration distributions. In the next phase of the ALARMS project, we expect to construct richer models of pilot state, based on Cognitive Work Analysis (CWA).

- **Rewards**: Reward is achieved for addressing the hazard and transitioning back to a nominal state. (Each hazard can have a different reward associated with it.) Actions with a low level of automation (hazards handled by the pilot) accumulate greater reward, whereas actions taken with a higher level of automation (handled by the system, without pilot feedback) achieve a lower level of reward. As actions that provide higher rewards usually take longer to execute, given a time deadline $\Delta$ after which no rewards can be earned, an optimal TMDP policy must often tradeoff high reward actions for their faster, but lower reward counterparts. We illustrate these trade-offs in the next Section, where optimal TMDP policies are found using the CPH algorithm (Marecki et al., 2007). In this work, we assume low levels of automation represent GUIs that provide more information to the pilot, which achieves higher reward. The assumption is limited to the reward and transition model, which is easily changed if desired.

## 3 Application

We constructed an analysis tool to allow the flight deck designer to understand the behavior of the flight deck for different hazard and pilot states. A picture of the tool can be seen in Figure 5. In the example shown, four sensor systems are considered (Weather Radar, Ice Protection, VNAV, and CDTI) which provide alerts. Options for the level of alert (Directive/Warning/Caution/Advisory/None) are configurable through an .xml file. The systems are run through the Bayesian Network to determine the underlying hazard state. Adjustable sliders allow the user to configure the rewards for addressing each hazard. (Since the TMDP reward function is defined using rewards for addressing the actual underlying hazards, instead of the sensor system rewards, the reward for the hazard is merely the maximum of the connected sensor system rewards). Phase of Flight can also be selected to determine the pilot state. Clicking the "Run" button induces the program to perform two actions: (1) It reads in the Bayes Net (in .xml format), and finds the underlying hazards by using the user-selected GUI options as evidence and (2) It builds and solves the underlying TMDP to find an optimal plan that addresses the hazards, and plots the results.

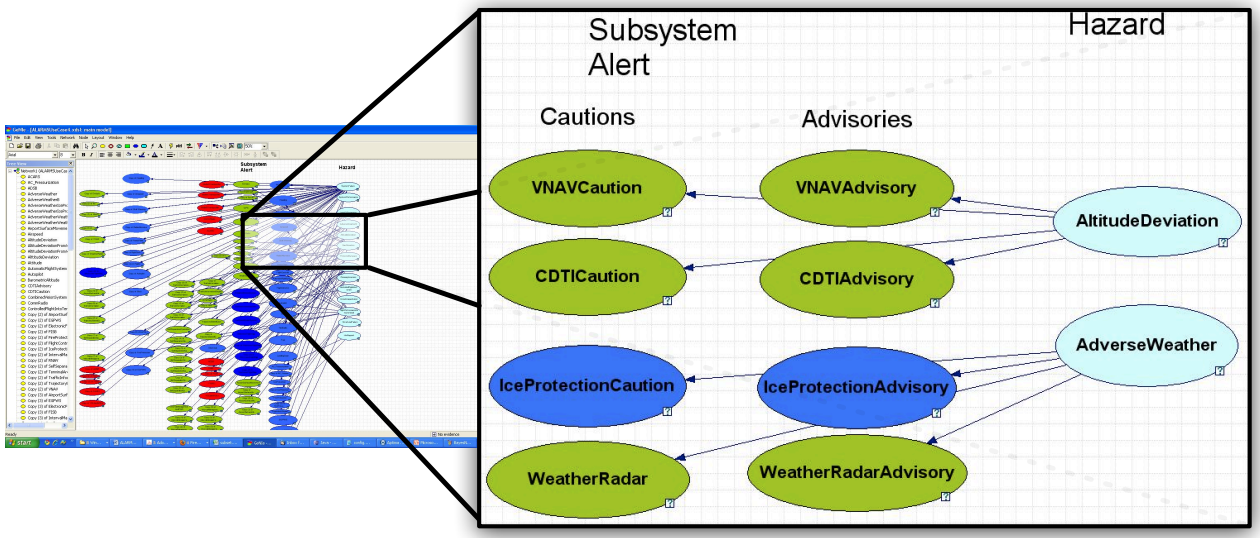

Figure 4: The ALARMS Bayesian Network.

For our example problems containing four sensors and two hazards, the Bayesian computation took less than a second, and the CPH solver found the optimal solution in $40.5$ seconds. For aircraft deployment, we anticipate producing the TMDP policies for hazard combinations in advance, and at flight time implementing the resulting policies through a lookup table.

Figure 5 displays the policies for various hazard and pilot states. The horizontal axes in all the plots mark time, and without loss of generality the graphs assume there is a deadline at the 20 second mark. The vertical axes represent the expected value (=sum of expected rewards). Differing actions are plotted in different colors (or can be seen as a break in the graph for those who read this in black and white). The action with the largest expected value for a given point in time should be executed if a decision is to be made at that point in time.

Consider the first plot in Figure 5, with two hazards to be addressed: hazard 1 = "Weather" and hazard 2 = "Altitude Deviation". The TMDP state considered in this plot is $(A, N)$ which means that the "Weather" hazard is in Advisory mode and the "Altitude Deviation" hazard is in Nominal mode. As can be seen, when the deadline is $> 5$ seconds away, action "L10" (shorthand for "Handle Hazard 1 at automation level 1, Handle Hazard 2 at automation level 0") is selected, the pilot is expected to handle the "Weather" hazard with some importance and the "Altitude Deviation" hazard with no importance. However, as we approach the deadline, action "L20" is more preferable, as automation level "2" is expected to act more quickly, thus potentially providing reward more quickly. (Note that only the upper envelope, that is the parts of the value functions that are not dominated by other value functions, are shown in the Figure.)

Subsequent plots in Figure 6 show how the optimal TMDP policy changes when we consider different states. The Figure 6a shows the optimal policy for a state $(C, A)$. Notice, that when the deadline is far away, both hazards are assigned to the pilot (automation level 1, the portion of the curve shown in green). However, as the deadline nears, the system deduces that there is not sufficient time for the pilot to handle both hazards. Consequently, the system assigns the less severe hazard ("Altitude Deviation") to the automation, shown as the purple portion of the curve on the right of the graph. Conversely in (Figure 6b), when we set a higher relative reward for addressing the "Altitude Deviation" hazard (seen on the number next to the slider bars on the right), as the deadline approaches, the system makes the opposite decision. It assigns the less prioritized (rewarded) hazard ("Weather") to the automation. On the right of the application, under "Actions", we can see that the purple right-portion of the plot now represents "L21" instead of "L12". Finally, the last plot in Figure 6 shows the effects of changing the phase of flight on the TMDP policy. By changing the phase of flight from "Enroute" to "Land", the assumptions about the pilot state have changed. During the Landing phase, the pilot is assumed to be less efficient than usual at performing tasks, and thus the cross-over point between actions is shifted left, reflecting the fact that the action with the higher level of automation is favored farther away from the deadline than in the graphs above it.

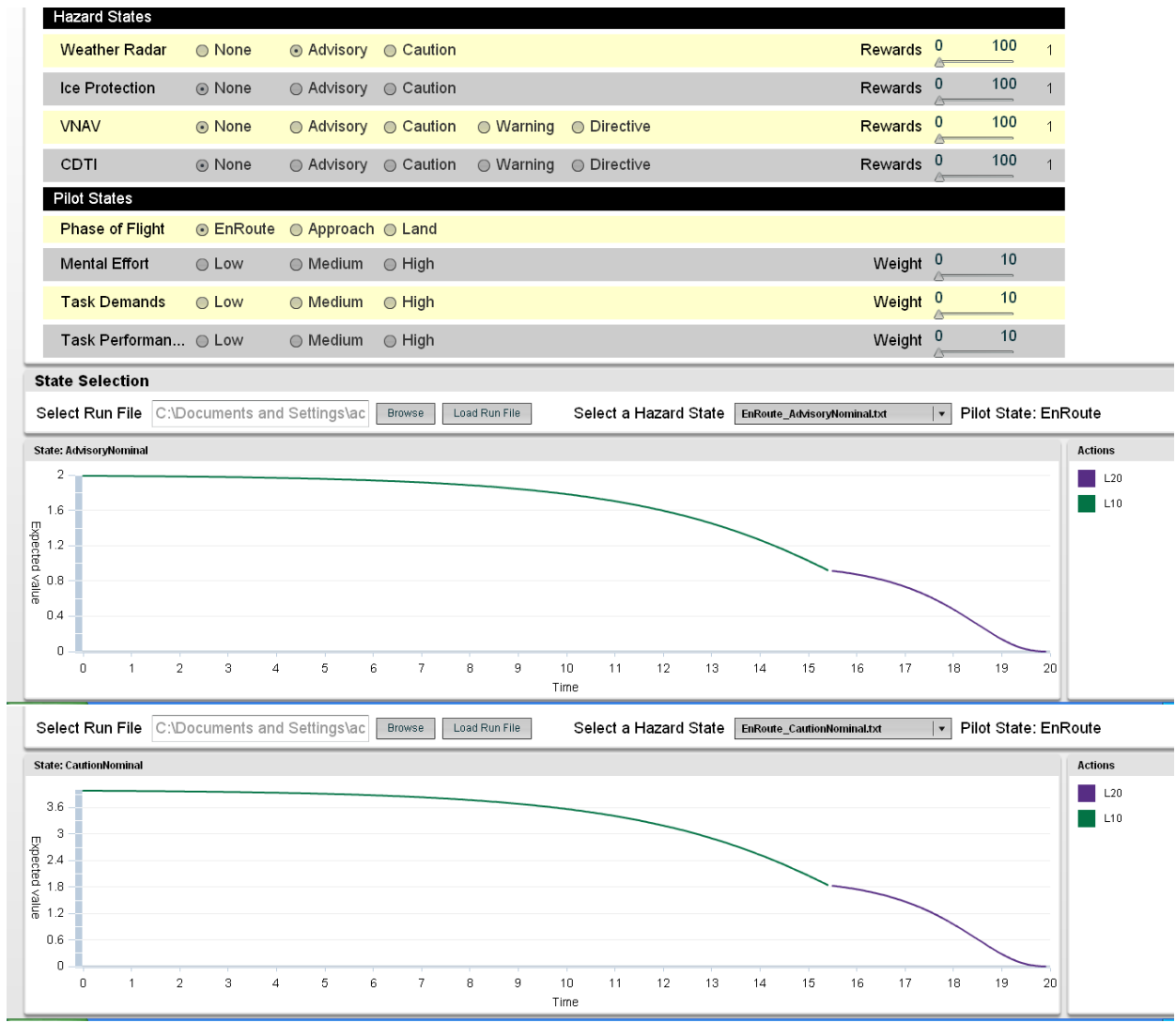

Figure 5: The ALARMS application. The top plot shows an Advisory for one hazard, the bottom plot was generated by changing the Advisory to a Caution (resulting in a doubling/rescaling of the y-axis).

## 4 Related Work

There are many works on the use of Bayesian networks for diagnosis (Andreassen et al., 1987; Breese et al., 2000). The idea of a time-dependent Markov model was first mentioned in (Boyan and Littman, 2000), and has been adapted for application (Rachelson et al., 2008). Progress towards solving these problems in fast time was made in recent publications (Feng et al., 2004; Li and Littman, 2005; Marecki et al., 2007).

Systems where MDPs or POMDPs were used for adjustable autonomy include (Scerri et al., 2002; Varakantham et al., 2005). These works did not allow for error checking or further consideration once an assignment was made. Furthermore, the state space was very large, not taking advantage of the TMDP framework, resulting in thousands of states for similarly sized problems.

As the cockpit has grown more complicated, numerous work has been published studying the effects on pilots. Cognitive Work Analysis has been used in order to model the effects of varying system states on pilot workload, and the effects of workload on performance (Vicente, 1999).

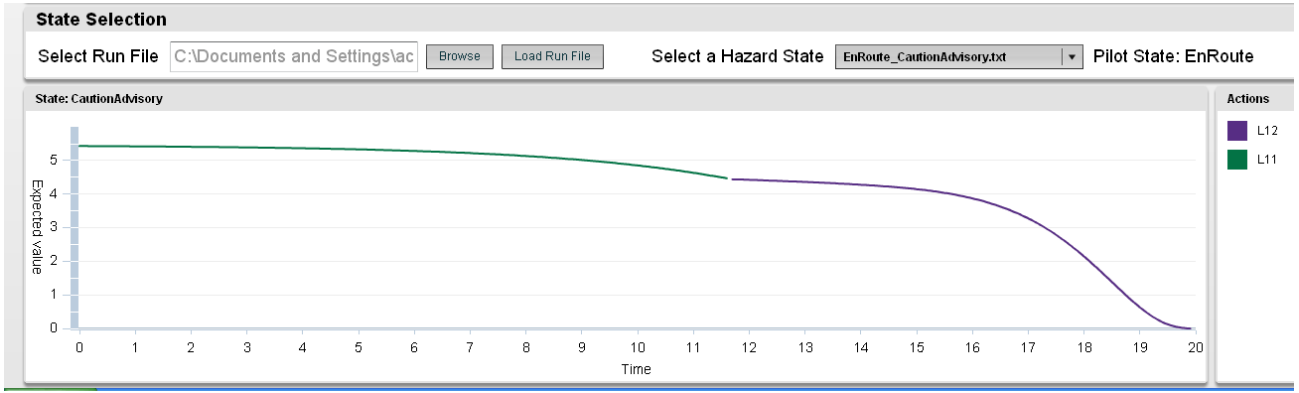
(a) Two hazards. The second hazard is handled at a higher level of automation, close to the deadline.

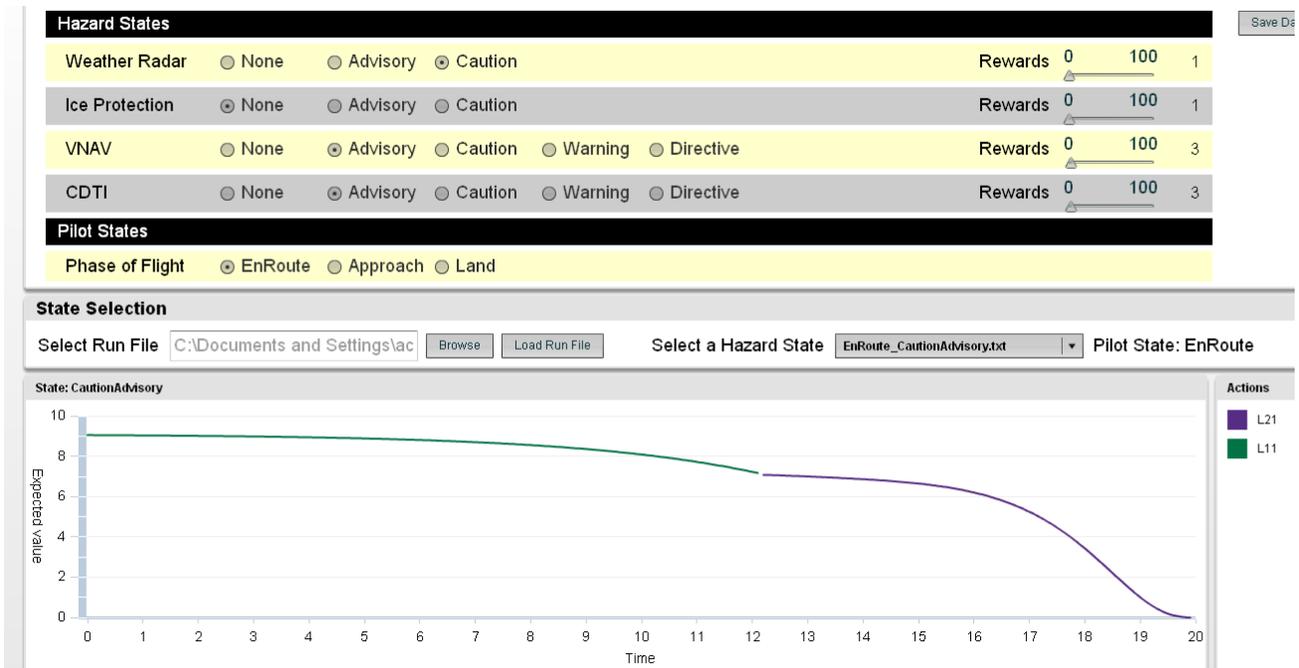
(b) The second hazard reward is increased via the values next to the slider bars, thus now it is handled by the pilot.

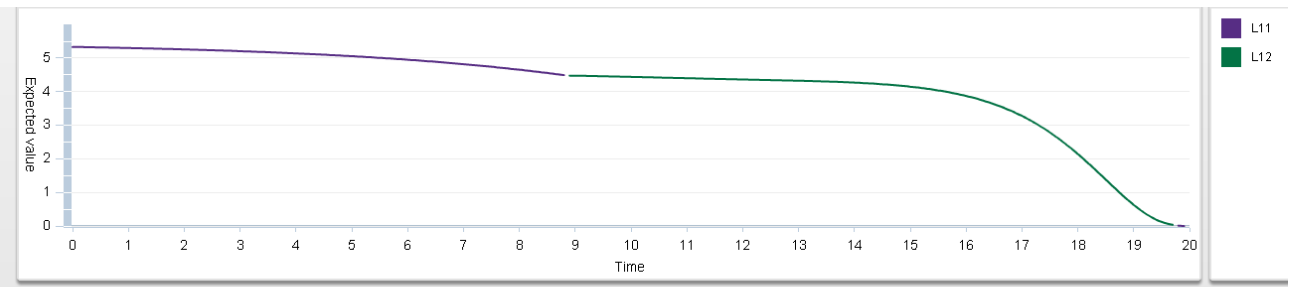
(c) Reward as in (a), but phase of flight is changed to land. The pilot's attention is diverted and the graph is shifted left as compared to (a).

Figure 6: The ALARMS application

Much work attempts to study the interaction between the human pilot and the automated system and include the levels of automation concept (Galster, 2003; Parasuraman et al., 2000). However, these works have not produced algorithms, whereby the assignment of tasks to pilot and human were varied automatically in a planned manner.

## 5 Conclusion

In the ALARMS project, we have developed several components necessary for operation of aircraft in a NextGen environment. First, a study of existing systems was conducted, and a matrix correlating aircraft sensor output (both legacy and NextGen) with real-world hazards was constructed. Second, this matrix was used to create a Bayesian Network whereby sensor output becomes evidence, and the presence and severity of real-world hazards is derived. Third, a TMDP model was created, allowing the aircraft sensor system to select the appropriate level of automation which best addresses hazards in a time-dependent environment. Finally, a demonstration application was created, linking the applications and thereby producing automation plans directly from the simulated sensor output.

Future work will continue in several directions. First, TMDP models will be scaled to handle not just the use cases from the demonstration application, but the whole Bayesian network. Second, work from the cognitive sciences literature will be leveraged to better estimate the pilot state. Third, we will consider the time-sensitive nature of the automated components, as in (Hansen and Zilberstein, 2001). Finally, we will develop a user interface based on levels of automation. Through the combined efforts of hazard state estimation, pilot state estimation, and human-automation planning, it is our hope to provide a robust, smooth transition to the Next Generation aircraft cockpit.

## 6 Acknowledgements

We would like to thank Gilbert Mizrahi for the graphical interface, and Amy Alexander for helpful guidance on pilot models.

This paper is based upon work supported by the National Aeronautics and Space Administration (NASA) under Contract No. NNL08AA20B issued through the Aviation Safety Program and monitored by Kara Latorella, whom the authors wish to thank. Any opinions, findings, and conclusions or recommendations expressed in this paper are those of the authors and do not necessarily reflect the views of NASA.